\documentclass[11pt,twoside]{article} 
\usepackage{acta-info}
\usepackage{euler}
\usepackage{multirow}
\usepackage{array}
\usepackage{booktabs}
\usepackage[section]{placeins}
\usepackage{float}
\usepackage{epstopdf}
\usepackage{graphicx}
\usepackage{caption}
\usepackage{subcaption}
\usepackage{algorithm}
\usepackage{algpseudocode}

\newcommand{\vol}{\textbf{x,} y (z) x--y}   
\newcommand{\amss}{\amssubj{%
		58A20
	}}     
	\newcommand{\CRs}{\CRsubj{%
			I.1.4
		}}   
		\newcommand{\keyw}{\keywords{%
				jet, cluster algorithm, hierarchical clustering, deep q-learning, neural network, multi-core, keras, cntk, louvain
			}}

\begin{document}
				
				\newcommand{\be}{\begin{equation}}
				\newcommand{\ee}{\end{equation}}
				\newcommand{\ba}{\begin{eqnarray}}
				\newcommand{\ea}{\end{eqnarray}}
				\newcommand{\nn}{\nonumber \\}
				

				\title{%
					Hierarchical clustering with deep Q-learning
				}
				
				\maketitle

				
				
				
				\twoauthors{%
					\href{http://people.inf.elte.hu/forceuse/}{Rich\'ard FORSTER}
				}{%
				\href{http://www.inf.elte.hu}{E\"otv\"os University} 
			}{%
			\href{mailto:forceuse@inf.elte.hu}{forceuse@inf.elte.hu} 
		}{%
		\href{http://compalg.inf.elte.hu/~fulop/}{\'Agnes F\"UL\"OP} 
	}{%
	\href{http://www.inf.elte.hu}{E\"otv\"os University}
}{%
\href{mailto:fulop@caesar.elte.hu}{fulop@caesar.elte.hu} 
}


\short{%
	R. Forster, \'A. F\"ul\"op
}{%
Clustering with deep learning 
}

\begin{abstract}
The reconstruction and analyzation of high energy particle physics data is just as important as the analyzation of the structure in real world networks. In a previous study it was explored how hierarchical clustering algorithms can be combined with  $k_t$ cluster algorithms to provide a more generic clusterization method. Building on that, this paper explores the possibilities to involve deep learning in the process of cluster computation, by applying reinforcement learning techniques.
The result is a model, that by learning on a modest dataset of $10,000$ nodes during $70$ epochs can reach $83,77\%$ precision in predicting the appropriate clusters.	
\end{abstract}

\section{Introduction} 
Different datasets should be clusterized with specific approaches. For real world networks, hierarchical algorithms, like the Louvain method, provides an efficient way to produce the clusters. Fusing some of the aspects of these processes and the $k_t$ jet clustering, a more generic process can be conceived as it was studied in \cite{ff-h}. This solution might prove very useful for heavy ion physics, where the jet physics plays an important role.
The contribution in this paper is a deep learning method, that uses reinforcement learning, to teach an artificial neural network how to clusterize the input graphs without any external user interaction.
The evaluation is provided on real world network data, that conforms the original Louvain method's properties, so a more thorough examination is possible.
Looking at the results, the neural network is capable to achieve an average precision on the test dataset of $83,77\%$, even by running for only $70$ epochs.

\section{Hierarchical clustering}
This section contains a brief introduction of the used hierarchical clustering algorithm and of the jet algorithms from physics.

\subsection{The Louvain algorithm}\label{louvain}
The Louvain method \cite{louvain}, is a multi-phase, iterative, greedy hierarchical clusterization algorithm, working on undirected, weighted graphs. The algorithm processes through multiple phases, within each phase multiple iterations until a convergence criteria is met. Its parallelization was explored in \cite{parlouvain}, that was further evolved into a GPU based implementation as was detailed in \cite{ines}. The modularity is a monotonically increasing function, spreading across multiple iterations, giving a numerical representation on the quality of the clusters. Because the modularity is monotonically increasing, the process is guaranteed to terminate. Running on a real world dataset, termination is achieved in not more than a dozen iterations.

\subsubsection{Modularity}\label{modularity}
On a set, $S={C_1,C_2,...,C_k}$, containing every community in a given partitioning of $V$, where $1\leq k \leq N$ and $V$ is the set of nodes, modularity $Q$ is given by the following \cite{findcom}:

\begin{equation}
\label{eq3}
Q=\frac{1}{2W}\sum\limits_{i\in V}e_{i\rightarrow C(i)}-\sum\limits_{C\in S}(\frac{deg_C}{2W}\cdot{}\frac{deg_C}{2W}),
\end{equation}

where $deg_C$ is the sum of the degrees of all the nodes in community C and $W$ is the sum of the weight of all the edges.

Modularity has multiple variants, like the ones described in \cite{rescom}, \cite{modgraph} and \cite{comdetres}. Yet the one defined in Eq. \ref{eq3} is the more commonly used.

\subsection{Jet algorithm}\label{jet}
During the last 40 years several jet reconstruction algorithms have been developed for hadronic colliders \cite{17c.}\cite{8.}. The first ever jet algorithm was published by Sterman and Weinberg in the 1970's \cite{1.}.
The cone algorithm plays an important role when a jet consists of a large amount of hadronic energy in a small angular region. It is based on a combination of particles with their neighbours in $\eta-\varphi$ space  within a cone of radius $R=\sqrt{(\Delta \varphi^2+\Delta \eta^2)}$. However the sequential recombination cluster algorithms combine the pairs of objects which have very close $p_t$ values. The particles merge into a new cluster through successive pair recombination. The starting point is the lowest $p_t$ particles for clustering in the $k_t$ algorithm, but in the anti-$k_t$ recombination algorithm it is the highest momentum particles. 

The jet clustering involves the reconstructed jet momentum of particles, which leaves the calorimeter together with modified values by the tracker system.

\subsubsection{Cone algorithm}
The Cone algorithm is one of the regularly used methods at the hadron colliders. The main steps of the iteration are the following \cite{1.}:
the seed particle $i$ belongs to the initial direction, and it is necessary to sum up the momenta of all particle $j$, which is situated in a circle of radius $R$ ($\Delta R_{ij}^2 = (y_i-y_j)^2+(\varphi_i-\varphi_j)^2< R^2$) around $i$, where $y_i$ and $\varphi_i$ are the rapidity and azimuth of particle $i$. 

The direction of the sum is applied as a new seed direction. The iteration procedure is repeated as long as the direction of the determined cone is stable.

It is worth noting what happens when two seed cone overlaps during the iteration. Two different groups of cone algorithms are discussed:
One possible solution is to select the first seed particle that has the greatest transverse momentum. Have to find the corresponding stable cone, i.e. jet and delete the particles from the event, which were included in the jet. Then choose a new seed, which is the hardest particle from the remaining particles, and apply to search the next jet. The procedure is repeated until there is no particle that has not worked. This method avoids overlapping.\\
Other possibility is the so called "overlapping" cones with the split-merge approach. All the stable cones are found, which are determined by iteration from all particles. This avoids the same particle from appearing in multiple cones. The split-merge procedure can be used to consider combining pair of cones. In this case more than a fraction $f$ of the transverse momentum of the softer cones derives from the harder particles; otherwise the common particles assigned to the cone, which is closer to them. The split-merge procedure applies the initial list of protojets, which contains the full list of stable cones:

\begin{enumerate}
	\item Take the protojet with the largest $p_t$ (i.e. hardest protojet), label it $a$.
	\item Search the next hardest protojet that shares particles with $a$ (i.e. overlaps), label it $b$. If no such protojet exists, delete $a$ from the list of protojets and add it to the list of final jets.
	\item Determine the total $p_t$ of the particles, which is shared between the two protojets, $p_{t,shared}$.
	\begin{itemize}
		\item If $p_{t,shared}>f$, where $f$ is a free parameter, it is called the overlap threshold, replace protojets $a$ and $b$ with a single merged protojet.
		\item Otherwise the protojets are scattered, for example assigning the shared particles to the protojet whose axis is closer.
	\end{itemize}
	\item Repeat from step 1 as long as there are protojets left.
\end{enumerate}

A similar procedure to split-merge method is the so called split-drop procedure, where the non-shared particles, which fall into the softer of two overlapping cones are dropped, i.e. are deleted from the jets altogether.

\subsubsection{Sequential recombination jet algorithm}
They go beyond just finding jets and implicitly assign a clustering sequence to an event, which is often closely connected with approximate probabilistic pictures that one may have for parton branching. The current work focuses on the $k_t$ algorithm, whose parallelisation was studied in \cite{13.} and \cite{ff-p}.

\paragraph{The $k_t$ algorithm for hadrons}

In the case of the proton-proton collision, the variables which are invariant under longitudinal boots are applied. These quantities which were introduced by \cite{2.} and the distance measures are longitudinally invariant as the following:
\begin{equation}
d_{ij}=min(p_{ti}^2,p_{tj}^2)\Delta R_{ij}^2, \;\;\;\;\; \Delta R_{ij}=(y_i-y_j)^2+(\varphi_i-\varphi_j)^2 \label{eq1} 
\end{equation}
\begin{equation}
d_{iB}=p_{ti}^2.\label{eq2}
\end{equation}

In this definition the two beam jets are not distinguished.

If $p=-1$, then it gives the  "anti-$k_t$" algorithm. In this case the clustering contains hard particles instead of soft particles. Therefore the jets extend outwards around hard seeds. Because the algorithm depends on the energy and angle through the distance measure, therefore the collinear branching will be collected at the beginning of the sequence.

\paragraph{Hierarchical $k_t$ clustering}\label{hierjet}
In \cite{ff-h} it was studied how to do hierarchical clustering, following the rules of the $k_t$ algorithm. First the list of particles has to be transformed into a graph, with the particles themselves appointed as nodes. The distance between the elements is a suitable selection for a weight to all edges between adjacent particles. But as it eventually leads up to $n*(n-1)/2$ links, where $n$ is the number of nodes, a better solution is to make connections between nearest neighbours and to the second to nearest. If the particle's nearest "neighbour" is the beam, it will be represented with an isolated node.
While the Louvain algorithm relies on modularity gain to drive the computation, the jet clustering variant doesn't have the modularity calculation, as it is known that the process will end, when all particles are assigned to a jet.

The result of this clustering will still be a dendogram, where the leafs will represent the jets.

\section{Basic artificial neural networks}\label{ann}
Since the beginning of the 1990s the  artificial neural network (ANN) methods are employed widely in the high energy physics for the jet reconstruction and track identification\cite{bd2}\cite{hk}. These methods are well-known in  offline and online data analysis also. 

Artificial neural networks are layered networks of artificial neurons (AN) in which  biological neurons are modelled. The underlying principle of operation is as follows,  each AN receives signals from another AN or from environment, gathers these and creates an output signal which  is forwarded to another AN or the environment. An ANN contains one input layer, one or more hidden layers and one output layer of ANs. Each AN in a layer is connected to the ANs in the next layer. There are such kind ANN configurations, where the feedback connections are introduced to the previous layers. 

\subsection{Architecture}
An artificial neuron is denoted by a set of input signals $(x_1,x_2,\dots x_n)$ from the environment or from another AN. A weight $w_i$ $(i=1,\dots n)$ is assigned to each input signal. If the value of weight is larger than zero then  the signal is excited, otherwise the signal is inhibited. AN assembles all input signals, determines a net signal and propagates an output signal.  

\subsubsection{Types of artificial networks}
Some features of neural systems which makes them the most distinct from the properties of conventional computing:

\begin{itemize}
	\item The associative recognition of complex structures
	\item Data may be non-complete, inconsistent or noisy
	\item The systems can train, i.e. they are able to learn and organize themselves
	\item The algorithm and hardware are parallel
\end{itemize}

There are many types of artificial neural networks. In the high energy partic\-le physics the so-called multi-layer perception (MLP) is the most widespread. Here a functional mapping from input $x_k$ to output $z_k$ values is realised with a function $f_{z_k}$:
$$    z_k=f_{z_k}\left( \sum_{j=1}^{m+1}w_{k_j}f_{y_j}\left( \sum_{i=1}^{n+1}v_{ji}x_i\right) \right),
$$
where $v_{ji}$ are the weights between the input layer and the hidden layer, and $w_{kj}$ are the weights between the hidden layer and the output layer. This type of ANN is called feed-forward multi-layer ANN. 

It can be extended into a layer of functional units. In this case   an activation function is implemented for the input layer. This ANN type is called functional link ANN. The output of this ANN is similar such as previously ANN, without it has additional layer, which contains $q$ functions  $h_l(x_1\dots x_n) (l=1\dots q)$.  The weights between the input layer and the functional layer are $u_{li}=1$, if $h_l$ depends on $x_i$, and $u_{li}=0$ otherwise. The output of this ANN is:
$$ z_k=f_{z_k}\left( \sum_{j=1}^{m+1}w_{kj}f_{y_j}\left( \sum_{l=1}^{q+1}v_{jl}h_l(x_1\dots x_n)\right)\right).$$
The functional link ANNs provides  better computational time and accuracy then the simple feed-forward multi-layer ANN.

\paragraph{Application in High-Energy Physics}
The first application, which was published in 1988,  discussed a recurrent ANN for tracking reconstruction \cite{bd1}.
A recurrent ANN was also used for tracking reconstruction in LEP experiment\cite{cp}.

An article published about 
a neural network method which was applied to find efficient mapping between certain observed hadronic kinematical variables and the quark-gluon identify. With this method it is able to separate gluon  from quark jets originating from the Monte-Carlo generated $e^+e^-$ events \cite{ll}.  A possible discrimination method  is presented by the combination of a neural network and QCD to separate the quark and gluon jet of $e^+e^-$ annihilation \cite{ic}.

The neural network clusterisation algorithm was applied for the ATLAS pixel detector to identify and split merged measurements created by multiple charged particles\cite{kjcl}.
The neural network based cluster reconstruction algorithm which can identify overlapping clusters and improves overall particle position reconstruction \cite{kes}. 

Artificial intelligence offers the potential to automate challenging data-processing tasks in collider physics. To establish its prospects, it was explored to what extent deep learning with convolutional neural networks can discriminate quark and gluon jets \cite{kpt}.

\section{Q-learning}\label{qlearn}
\begin{figure}[h]
	\centering
	\includegraphics[width=3.5in]{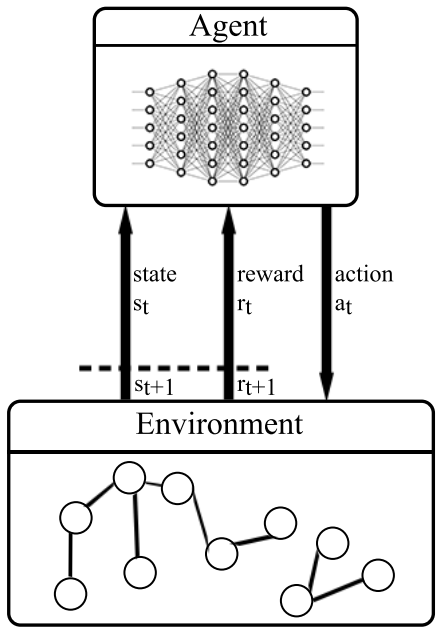}
	\caption{The agent-environment interaction in reinforcement learning}
	\label{agent_env}
\end{figure}

Q-learning is a model-free reinforcement learning technique \cite{rl}. The reinforcement learning problem is meant to be learning from interactions to achieve a goal. The learner and decision-maker is called the agent. The thing it interacts with is called the environment, that contains everything from the world surrounding the agent. There's a continuous interaction between, where the agent selects an action and the environment responds by presenting new situations (states) to the agent. The environment also returns rewards, special numerical values that the agent tries to maximize over time. A full specification of an environment defines a task, that is an instance of the reinforcement learning problem. 
Specifically, the agent and environment interact at each of a sequence of discrete time steps $t=0,1,2,\dots$. At each time step $t$, the agent receives the environment's state, $s_t \in S$, where $S$ is the set of possible states, and based on that it selects an action, $a_t \in \mathfrak{A}(s_t)$, where $\mathfrak{A}(s_t)$  is the set of all available actions in state $s_t$. At the next time step as a response to the action, the agent receives a numerical reward, $r_{t+1} \in \mathfrak{R}$ , and finds itself in a new state, $s_{t+1}$ (Figure \ref{agent_env}).

At every time step, the agent implements a mapping from states to probabilities of selecting the available actions. This is called the agent's policy and is denoted by $\pi_t$, where $\pi_t(s,a)$ is the probability that $a_t = a$ if $s_t = s$. Reinforcement learning methods specify how the agent changes this using its experience. The agent's goal is to maximize the total amount of reward it receives over the long run.

\subsection{Goals and rewards}\label{reward}
The purpose or goal of the agent is formalized in terms of a special reward passed from the environment. At each time step, the reward is a simple number, $r_t \in \mathfrak{R}$. The agent's goal is to maximize the total  reward it receives.

\subsection{Returns}\label{return}
If the rewards accumulated after time step $t$ is denoted by $r_{t+1},r_{t+2},r_{t+3},\dots$, what will be maximized by the agent is the expected return $R_t$, that is defined as some function of the received rewards. The simplest case is the sum of the rewards: $R_t=r_{t+1}+r_{t+2}+r_{t+3}+\dots+r_T$, where $T$ is the final time step.
This approach comes naturally, when the agent-environment interaction breaks into subsequences, or episodes. Each episode ends in a special terminal state, that is then being reset to a standard starting state. The set of all nonterminal states is denoted by $S$, while the set with a terminal state is denoted by $S^+$.

Introducing discounting, the agent tries to maximize the the sum of the discounted rewards by selecting the right actions. At time step $t$ choosing action $a_t$, the discounted return will be defined with equation \ref{3.2}. 
\begin{equation}
R_t=r_{t+1}+\gamma r_{t+2}+\gamma^2 r_{t+3}+\dots=\sum_{k=0}^{\infty}\gamma^k r_{t+k+1},
\label{3.2}
\end{equation}
where $\gamma$ is a parameter, $0\leq\gamma\leq1$, called the discount rate. 
It determines the present value of future rewards: a reward received at time step $t+k$ is worth only $\gamma^{k-1}$ times the immediate reward. If $\gamma < 1$, the infinite sum still is a  finite value as long as the reward sequence $\{r_k\}$ is bounded. If $\gamma=0$, the agent is concerned only with maximizing immediate rewards. If all actions influences only the immediate reward, then the agent could maximize equation \ref{3.2} by separately maximizing each reward. In general, this can reduce access to future rewards and the return may get reduced. As $\gamma$ approaches 1, future rewards are used more strongly.

\subsection{The Markov property}\label{markov}
Assuming a finite set of states and reward values, also considering how a general environment responds at time $t+1$ to the action taken at time $t$, this response may depend on everything that has happened earlier. In this case only the complete probability distribution can define the dynamics:

\begin{equation}
Pr\{s_{t+1}=s',r_{t+1}=r|s_t,a_t,r_t,s_{t-1},a_{t-1},\dots,r_1,s_0,a_0\},
\label{3.4}
\end{equation}

for all $s'$,$r$, and all possible values of the past events: $s_t,a_t,r_t,\dots,r_1,s_0,a_0$. If the state has the Markov property the environment's response at $t+1$ depends only on the state and action at $t$ and the dynamics can be defined by applying only equation \ref{3.5}.  

\begin{equation}
Pr\{s_{t+1}=s',r_{t+1}=r|s_t,a_t\},
\label{3.5}
\end{equation}

for all $s'$, $r$, $s_t$, and $a_t$. Consequently if a state has the Markov property, then it's a Markov state, only if \ref{3.5} is equal to \ref{3.4} for all $s'$, $r$, and histories, $s_t,a_t,r_t,\dots,r_1,s_0,a_0$. In this case, the environment has the Markov property.

\subsection{Markov cecision process}\label{MDP}
A reinforcement learning task satisfying the Markov property is a Markov decision process, or MDP. If the state and action spaces are finite, then it is a finite MDP. This is defined by its state and action sets and by the environment's one-step dynamics. Given any state, action pair, $(s,a)$, the probability of each possible next state, $s'$, is  

\[
P_{ss'}^a=Pr\{s_{t+1}=s'|s_t=s,a_t=a\}
\label{3.6}
\]

Having the current state and action, $s$ and $a$, with any next state, $s'$, the expected value of the next reward can be computed with  

\[
R_{ss'}^a=E\{r_{t+1}|s_t=s,a_t=a,s_{t+1}=s'\}.
\label{3.7}
\]

These quantities, $P_{ss'}^a$ and $R_{ss'}^a$, completely specify the most important aspects of the dynamics of a finite MDP.

\subsection{Value functions}
Reinforcement learning algorithms are generally based on estimating value functions, that are either functions of states or state-action. They estimate how good a given state is, or how good a given action in the present state is. How good it is, depends on future rewards that can be expected, more precisely, on the expected return. As the rewards received depends on the taken actions, the value functions are defined with respect to particular policies. 
A policy, $\pi$, is a mapping from each state, $s\in S$, and action, $a\in A(s)$, to the probability $\pi(s,a)$ of taking action $a$ while in state $s$. The value of a state $s$ under a policy $\pi$, denoted by $V^\pi(s)$, is the expected return when starting in $s$ and following $\pi$. For MDPs $V^\pi(s)$ is defined as  

\[
V^\pi(s)=E_\pi\{R_t|s_t=s\}=E_\pi\{\sum_{k=0}^{\infty}\gamma^k r_{t+k+1}|s_t=s\},
\label{3.8}
\]

where $E_\pi$ is the expected value given that the agent follows policy $\pi$. The value of the terminal state is always zero. $V^\pi$ is the state-value function for policy $\pi$. Similarly, the value of taking action $a$ in state $s$ under a policy $\pi$, denoted by $Q^\pi(s,a)$ is defined as the expected return starting from $s$, taking the action $a$, and following policy $\pi$:  

\[
Q^\pi(s,a)=E_\pi\{R_t|s_t=s,a_t=a\}=E_\pi\{\sum_{k=0}^{\infty}\gamma^k r_{t+k+1}|s_t=s,a_t=a\}.
\label{3.9}
\]

$Q^\pi$ is the action-value function for policy $\pi$.

$V^\pi$ and $Q^\pi$ can be estimated from experience. If an agent follows policy $\pi$ and maintains an average of the actual return values in each encountered state, then it will converge to the state's value, $V^\pi(s)$, as the number of times that state is encountered approaches infinity. If in a given state every action has a separate average, then these will also converge to the action values, $Q^\pi(s,a)$.

\subsection{Optimal value functions}
To solve a reinforcement learning task, a specific policy needs to be found, that achieves a lot of reward over the long run. For finite MDPs, an optimal policy can be defined. Value functions define a partial ordering over policies. A policy $\pi$ is defined to be better than or equal to a policy $\pi'$ if its expected return is greater than or equal to $\pi'$ for all states. Formally, $\pi \geq \pi'$ if and only if $V^\pi(s)\geq V^{\pi'}(s)$ for all $s\in S$. At least one policy exists, that is better than or equal to all other policies and this is the optimal policy. If more than one exists, the optimal policies are denoted by $\pi^*$. The state-value function among them is the same, called the optimal state-value function, denoted by $V^*$, and defined as  

\[
V^*(s)=\max_\pi V^\pi(s),
\label{3.11}
\]

for all $s\in S$. 
The optimal action-value functions are also shared, denoted by $Q^*$, and defined as  

\[
Q^*(s,a)=\max_\pi Q^\pi(s,a),
\label{3.12}
\]

for all $s\in S$ and $a\in A(s)$. For the state-action pair $(s,a)$, this gives the expected return for taking action $a$ in state $s$ and following an optimal policy. Thus, $Q^*$ can be defined in terms of $V^*$ as follows:  

\[
Q^*(s,a)=E\{r_{t+1}+\gamma V^*(s_{t+1})|s_t=s,a_t=a\}.
\label{3.13}
\]

\section{Clustering with deep Q-learning}\label{deepq}
The Deep Q-learning (DQL) \cite{drl1} \cite{drl2} is about using deep learning techniques on the standard Q-learning (section \ref{qlearn}).

Calculating the Q state-action values using deep learning can be achieved by applying the following extensions to standard reinforcement learning problems:
\begin{enumerate}
\item Calculate Q for all possible actions in state $s_t$,
\item Make prediction for Q on the new state $s_{t+1}$ and find the action $a_{t+1}=\max_a a\in A(s_{t+1})$, that will yield the biggest return,
\item Set the Q return for the selected action to $r + \gamma Q(s_{t+1}, a_{t+1})$. For all other actions the return should remain unchanged,
\item Update the network using back-propagation and mini-batches stochastic gradient descent.
\end{enumerate}

This approach in itself leads to some additional problems. The {\em exploration-exploitation issue} is related to which action is taken in a given state. By selecting an action that always seems to maximize the discounted future reward, the agent is acting greedy and might miss other actions, that can yield higher overall reward in the long run. To be able to find the optimal policy the agent needs to take some exploratory steps at specific time steps. This is solved by applying the {\em $\epsilon-greedy$ algorithm} \cite{rl}, where a small probability $\epsilon$ will choose a completely random action.

The other issue is the problem of the {\em local-minima} \cite{locmin}. During training multiple states can be explored, that are highly correlated and this may make the network to learn replaying the same episode. This can be solved, by first storing past observations in a {\em replay memory} and taking random samples from there for the mini-batch, that is used to replay the experience.

\subsection{Environment}\label{env}
The environment provides the state that the agent will react to. In case of clustering the environment will be the full input graph. The actual state the necessary information required to compute the Louvain method,  packaged into a Numpy stack. These include the weights, degrees, number of loops, the actual community and the total weight of the graph. Each state represents one node of the graph with all of its neighbors. The returned rewards for each state will be based on the result of the actual Louvain clusterization, which means during training the environment will compute the real clusters. If the action selected by the agent leads to the best community, that will have a positive reward set to $10000$ and in any other case the returned value will be $-1000$. After stepping, the next state will contain the modified community informations.

The agent's action space is finite and predefined and the environment also has to reflect this. Let the cardinality of the action space be noted for all $s\in S$ states by $|A(s)|$ For this reason, the state of the environment contains information about only $|A(s)|$ neighbors. This can lead to more nodes, than how many really is connected to a given element. In this case the additional dummy node's values are filled with extremals, in the current implementation with negative numbers. One limitation of the actual solution is that if the number of neighbors are higher, than $|A(s)|$, then only the first $|A(s)|$ neighbors will be considered, in the order in which they appear in dataset.
The first "neighbor" will be currently evaluated node, so in case the clusterization will not yield any better community, the model should see, that the node stays in place.

To help avoid potential overflow during the computation, weights of the input graph are normalized to be between $0.000001$ and $1$.

\subsection{Agent}\label{agent}
The agent acts as the decision maker, selecting the next community for a given node. It takes the state of the environment as an input and gives back the index of the neighbor that is considered to be providing the best community.

\subsubsection{Implementation in Keras}\label{keras}
Keras \cite{keras} is a Python based high-level neural networks API, compatible with the TensorFlow, CNTK, and Theano machine learning frameworks. This API encourages experimentation as it supports rapid development of neural networks. It allows easy and fast prototyping, with a user friendly, modular, and extensible structure. Both {\em convolutional networks} and {\em recurrent networks} can be developed, also their combinations are also possible in the same agent. As all modern neural network API it both runs on CPU and GPU for higher performance.

The core data structure is a model, that is a collection of layers. The simplest type is the {\em Sequential model}, a linear stack of layers. More complex architectures also can be achieved using the Keras functional API.

The clustering agent utilizes a Sequential model:
\begin{verbatim}
from keras.models import Sequential

model = Sequential()
\end{verbatim}
Stacking layers into a model is done through the {\em add} function:
\begin{verbatim}
from keras.layers import Dense

model.add(Dense(128, input_shape=(self.state_size,
            self.action_size), activation='relu'))
model.add(Dropout(0.5))
model.add(Dense(128, activation='relu'))
model.add(Dropout(0.5))
model.add(Dense(128, activation='relu'))
\end{verbatim}

The first layer will handle the input and has a mandatory parameter defining its size. In this case {\em input\_shape} is provided as a 2-dimensional matrix, where {\em state\_size} is the number of parameters stored in the state and {\em action\_size} is the number of possible actions.
The first parameter tells how big the output dimension will be, so in this case the input will be propagated into a 128-dimensional output.

The following two layers are hidden layers (section \ref{ann}) with $128$ internal nodes, with {\em rectified linear unit} (ReLU) activation. The rectifier is an activation function given by the positive part of its argument: $f(x)=x^+=\max(0,x)$, where $x$ is the input to a neuron. The rectifier was first introduced to a dynamical network in \cite{rect}. It has been demonstrated in \cite{rectproof} to enable better training of deeper networks, compared to the widely used activation function prior 2011, the logistic sigmoid \cite{sigmoid}.

During training overfitting happens, when the ANN goes to memorize the training patterns. In this case the network is weak in generalizing on new datasets. This appears for example, when an ANN is very large, namely it has too many hidden nodes and hence, there are too many weights which need to be optimized.\\
The dropout for the hidden layers is used to prevent overfitting on the learning dataset. Dropout is a technique that makes some randomly selected neurons ignored during training. Their contribution to the activation of neurons on deeper layers is removed temporally and the weight updates are not applied back to the neurons. If neurons are randomly dropped during training, then others will have to handle the representation, that is required to make predictions, that is normally handled by the dropped elements. This results in multiple independent internal representations for the given features \cite{dropout}. This way the network becomes capable of better generalization and avoids potential overfitting on the training data.

The output so far will still be a matrix with the same shape as the input. This is flatten into a 1-dimensional array by adding the following layer:
\begin{verbatim}
model.add(Flatten())
\end{verbatim}

Finally to have the output provide the returns on each available actions, the last layer changes the output dimension to {\em action\_size}:
\begin{verbatim}
model.add(Dense(self.action_size, activation='linear'))
\end{verbatim}

Once the model is set up, the learning process can be configured with the {\em compile} function:
\begin{verbatim}
model.compile(loss='mse', optimizer=Adam(lr=self.learning_rate)),
\end{verbatim}

where {\em learning\_rate} has been set to $0.001$. For the loss function {\em mean squared error} is used, optimizer is an instance of {\em Adam} \cite{adam} with the mentioned learning rate. The discount rate for future rewards have been set to $\gamma=0.001$. This way the model will try to select actions, that yield the maximum rewards in the short term. While maximizing the reward in long term can eventually lead to a policy, that computes the communities correctly, choosing it this small makes the model learn to select the correct neighbors faster.

To make a prediction on the current state, the {\em predict} function is used:
\begin{verbatim}
model.predict(state.reshape(1, self.state_size,
                             self.action_size))
\end{verbatim}

For Keras to work on the input state, it always have to be reshaped into dimensions $(1, state\_size, action\_size)$, while the change always has to keep the same number of state elements.

\section{Results}\label{results}
Evaluation of the proposed solution is done by processing network clustering on undirected, weighted graphs. These graphs contain real network information, instead of evaluating on physics related datasets (section \ref{jet}), as it is more suitable for the original Louvain method. Because of this, the modularity can be used as a sort of metric to measure the quality (subsection \ref{louvain}) of the results. Additionally the number of correct predictions and misses are used to describe the deep Q-learning (section \ref{deepq}) based method's efficiency.

Numerical evaluations are done by generating one iteration on the first level of the dendogram as the top level takes the most time to generate as it is based on all the original input nodes. The GPU implementation of the Louvain method being used was first described in \cite{ines}.

\subsection{Dataset}
The proposed model, as well as the Louvain clustering works on undirected, weighted graphs. Such graphs can be generated from U.S. Census 2010 and Tiger/Line 
2010 shapefiles, that are freely available from \cite{census}. They contain the following:

\begin{itemize}
\item the vertices are the Census Blocks;                                 
\item there's an edge between two vertices if the corresponding Census Blocks share a line segment on their border
\item each vertex has two weights:  
\begin{itemize}
	\item Census2010 POP100 or the number of people living in that Census Block
	\item Land Area of the Census Block in square meters
\end{itemize}
\item the edge weights are the pseudo-length of the shared borderlines.
\item each Census Block is identified by a point, that is given longitudinal and latitudinal coordinates
\end{itemize}

A census block is the smallest geographical unit used by the United States Census Bureau for tabulation of 100-percent data. The pseudo-length is given by $\sqrt{(x^2 + y^2)}$, where $x$ and $y$ are the differences in longitudes and latitudes of each line segment on the shared borderlines. The final result is multiplied by $10^7$ to make the edge weights integers. For clusterization the node weights are not used.

The matrices used for evaluation contains the information related to New York, Oregon and Texas (table \ref{dataset}), that was arbitrary selected from the SuiteSparse Matrix Collection \cite{suitesparse}. The graph details can be found in \cite{sparse}.

\begin{table}[h]
	\centering
	\begin{tabular}{|c|c|c|c|}\hline
	& New York & Oregon & Texas\\ \hline
	Nodes & $350,169$ & $196,621$ & $914,231$\\
	\hline
	Edges & $1,709,544$ & $979,512$ & $4,456,272$\\
	\hline
	\end{tabular}
	\caption{Details of the selected datasets}
	\label{dataset}
\end{table}

Due to the limitations of the proposed solution as was described in subsection \ref{env}, $4$ neighbors are kept for each nodes during the computation.

\subsection{Precision of the neural network}
The model described in section \ref{deepq} have been trained on the Oregon graph, taking the first $10000$ nodes based on the order how they are first mentioned in the original dataset, running for $70$ epochs. The ratio of the good and bad predictions are shown in table \ref{prec_fig}.

\begin{table}[h]
	\centering
	\begin{tabular}{|c|c|c|c|}\hline
		& New York & Oregon & Texas\\ \hline
		Positive & $310,972$ & $172,028$ & $750,563$\\
		\hline
		Negative & $39,197$ & $24,593$ & $163,668$\\
		\hline
	\end{tabular}
	\caption{Precision of the deep learning solution}
	\label{prec_fig}
\end{table}

The deep learning solution's precision in average is $83,77\%$. Specifically on the datasets it's respectively $87,4\%$, $85,7\%$ and $78,2\%$. Precision can be further increased by running the training for more epochs or by further tune the hyperparameters.

\subsection{Modularity comparison}
The Louvain method assumes nothing of the input graph. The clusterization can be done without any prior information of the groups being present in the network. The modularity is presented (table \ref{mod_fig}) for all $3$ test matrices for both the Louvain algorithm and the deep Q-learning based solution.

\begin{table}[h]
	\centering
	\begin{tabular}{|c|c|c|c|}\hline
		& New York & Oregon & Texas\\ \hline
		DQL & $38,752.84$ & $6,659.93$ & $99,789.88$\\
		\hline
		Louvain & $45,339.84$ & $7,724.99$ & $120,745.75$\\
		\hline
	\end{tabular}
	\caption{Precision of the deep learning solution}
	\label{mod_fig}
\end{table}

The modularities showing similar results to the precision of the network: the New York graph  has a modularity less with $14,53\%$ compared to the Louvain computation, while Oregon is less with $13,8\%$ and Texas is less with $17,36\%$. This proves, that by loosing from the precision, the qualities of the clusters do not degrade more than, what is lost on the precision.

\section{Summary}
In this paper a new hierarchical clustering was proposed based on the Louvain method, using deep learning. The detailed model was capable to achieve $83,77\%$ of precision, while only being teached for $70$ epochs. Even with the error, the resulting modularity in average was less compared to the Louvain method's result with only $15,23\%$.

\section{Future work}
The current solution can't process the whole graph, just a subset of the neighbors are considered, when processing the communities. This needs to be extended further, to have a fully a realized deep learning clusterizer. The model still needs to be evaluated on Jet related datasets and needs to be explored if any changes are required in the agent to work efficiently on those type of graphs. Most of the errors come from choosing a dummy node as the best community, which implies that the way, how these nodes are represented should be studied further.

\end{document}